%% file: KDD.tex
\documentclass[sigconf]{acmart}

\AtBeginDocument{%
  }

\copyrightyear{2025}
\acmYear{2025}
\setcopyright{acmlicensed}\acmConference[KDD '25]{Proceedings of the 31st ACM SIGKDD Conference on Knowledge Discovery and Data Mining V.1}{August 3--7, 2025}{Toronto, ON, Canada}
\acmBooktitle{Proceedings of the 31st ACM SIGKDD Conference on Knowledge Discovery and Data Mining V.1 (KDD '25), August 3--7, 2025, Toronto, ON, Canada}
\acmDOI{10.1145/3690624.3709208}
\acmISBN{979-8-4007-1245-6/25/08}

\usepackage{booktabs}
\usepackage{amsfonts}
\usepackage{subfigure}
\usepackage{multirow}
\usepackage{amsmath}
\usepackage{mathrsfs}
\usepackage{mathtools}
\usepackage{amsthm}
\usepackage{algorithm}
\usepackage{algorithmic}
\usepackage[normalem]{ulem}
\useunder{\uline}{\ul}{}

\theoremstyle{plain}
\newtheorem{theorem}{Theorem}[section]

\newtheorem{lemma}[theorem]{Lemma}
\newtheorem{corollary}[theorem]{Corollary}
\theoremstyle{definition}

\theoremstyle{remark}

\begin{document}

\title{Enhancing Unsupervised Graph Few-shot Learning via Set Functions and Optimal Transport}

\author{Yonghao Liu}
\authornote{Key Laboratory of Symbolic Computation and Knowledge Engineering of the Ministry of Education}
\affiliation{
  \institution{College of Computer Science and Technology, Jilin University}
  \city{Changchun}
  \country{China}
}
\email{yonghao20@mails.jlu.edu.cn}

\author{Fausto Giunchiglia}
\affiliation{%
  \institution{Department of Information Engineering and Computer Science, University of Trento}
  \city{Trento}
  \country{Italy}
}
\email{fausto.giunchiglia@unitn.it}

\author{Ximing Li}
\authornotemark[1]
\affiliation{%
  \institution{College of Computer Science and Technology, Jilin University}
  \city{Changchun}
  \country{China}
}
\email{liximing86@gmail.com}

\author{Lan Huang}
\authornotemark[1]
\affiliation{%
  \institution{College of Computer Science and Technology, Jilin University}
  \city{Changchun}
  \country{China}
}
\email{huanglan@jlu.edu.cn}

\author{Xiaoyue Feng}
\authornotemark[1]
\authornote{Corresponding author}
\affiliation{%
  \institution{College of Computer Science and Technology, Jilin University}
  \city{Changchun}
  \country{China}
}
\email{fengxy@jlu.edu.cn}

\author{Renchu Guan}
\authornotemark[1]
\authornotemark[2]
\affiliation{%
  \institution{College of Computer Science and Technology, Jilin University}
  \city{Changchun}
  \country{China}
}
\email{guanrenchu@jlu.edu.cn}

\renewcommand{\shortauthors}{Yonghao Liu et al.}
\begin{abstract}
Graph few-shot learning has garnered significant attention for its ability to rapidly adapt to downstream tasks with limited labeled data, sparking considerable interest among researchers. Recent advancements in graph few-shot learning models have exhibited superior performance across diverse applications. Despite their successes, several limitations still exist. First, existing models in the meta-training phase predominantly focus on instance-level features within tasks, neglecting crucial set-level features essential for distinguishing between different categories. Second, these models often utilize query sets directly on classifiers trained with support sets containing only a few labeled examples, overlooking potential distribution shifts between these sets and leading to suboptimal performance. Finally, previous models typically require  necessitate abundant labeled data from base classes to extract transferable knowledge, which is typically infeasible in real-world scenarios. 
To address these issues, we propose a novel model named \textbf{STAR}, which leverages \textbf{S}et func\textbf{T}ions and optim\textbf{A}l t\textbf{R}ansport for enhancing unsupervised graph few-shot learning. Specifically, STAR utilizes expressive set functions to obtain set-level features in an unsupervised manner and employs optimal transport principles to align the distributions of support and query sets, thereby mitigating distribution shift effects. Theoretical analysis demonstrates that STAR can capture more task-relevant information and enhance generalization capabilities. Empirically, extensive experiments across multiple datasets validate the effectiveness of STAR. Our code can be found \href{https://github.com/KEAML-JLU/STAR}{\textcolor{red}{here}}.
\end{abstract}



\maketitle
\input{introduction}
\input{related_work}
\input{preliminary}
\input{method}
\input{theory}
\input{experiment}
\input{result}

\input{conclusion}

\section*{Acknowledgments}
This work is supported in part by funds from the National Key Research and Development Program of China (No. 2021YFF1201200), the National Natural Science Foundation of China (No. 62172187 and No. 62372209). Fausto Giunchiglia’s work is funded by European Union’s Horizon 2020 FET Proactive Project (No. 823783).
\bibliographystyle{ACM-Reference-Format}
\balance
\bibliography{sample-base}

\appendix
\input{appendix}

\end{document}

%% file: introduction.tex
\section{Introduction}
Graphs, as a fundamental data structure, effectively capture relationships between objects in diverse real-world systems. Graph neural networks (GNNs) demonstrate impressive expressive capabilities for handling graph-structured data \cite{kipf2016semi, liu2021deep, liu2023time, liulocal, liu2024simple, liu2024improved, liu2024resolving, liu2025improved, liu2025boosting}. However, prevailing GNN models depend heavily on a substantial number of labeled nodes to achieve optimal performance \cite{li2024simple, liu2024simple}. Hence, graph few-shot learning (FSL) models \cite{zhou2019meta, huang2020graph}, which aim to learn generalizable meta-knowledge from a limited number of labeled nodes from source tasks to swiftly adapt to unseen target tasks, are undergoing vigorous development. Generally, these models adhere to a two-stage learning paradigm. In the meta-training stage, a meta-learner is trained on abundant diverse tasks. Next, in the meta-testing stage, the meta-learner is fine-tuned with few labeled data and then applied to unlabeled ones. Although previous graph FSL models achieve notable success, they still have several limitations. 

\textbf{First}, during meta-training, they mainly focus on learning meaningful instance-level features within each task to train the meta-learner. It is apparent that the significance of instance-level features for generalization to novel tasks, which can enhance the model's adaptability to downstream tasks \cite{lee2023self}. However, graph FSL models are fundamentally trained based on individual tasks as the basic training unit, rather than individual data, as in traditional deep learning paradigms. They learn from different but related tasks and can also be regarded as set-level problems, where the input set corresponds to the training data for each task \cite{lee2023self, lee2022set}. Moreover, the order of the elements within the set has no impact on the model's output. Therefore, permutation-invariant set-level features within tasks should be taken into consideration. Several works \cite{gordon2018meta, triantafillou2019meta} in the Euclidean domain, such as image analysis, have proven that deriving discriminative set representations from a few samples is critical for facilitating downstream tasks. Nevertheless, existing graph FSL models for non-Euclidean graph-structured data have not explored the potential benefits brought by set-level features and lack targeted algorithm designs. 
\begin{figure}
    \centering
    \includegraphics[width=0.36\textwidth]{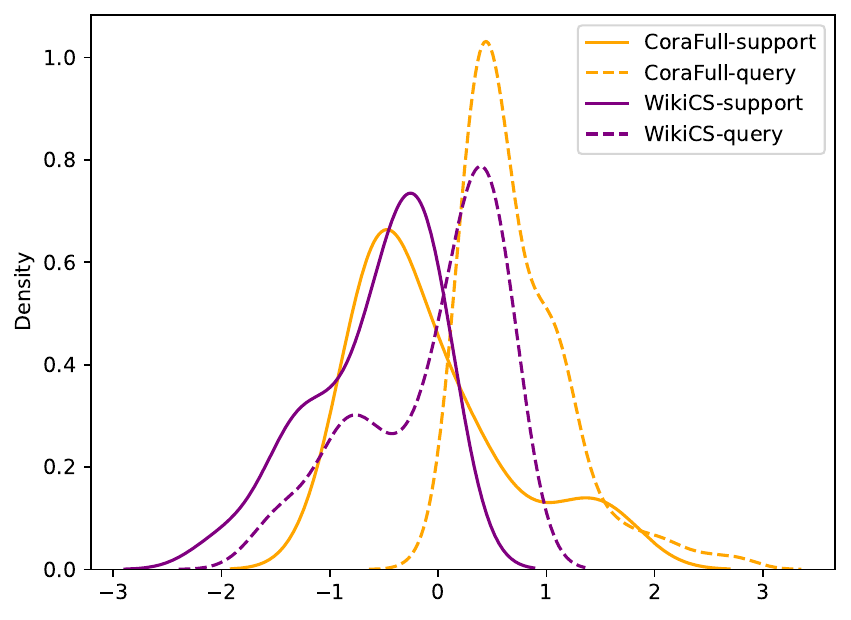}
    \caption{Data Distributions of support and query sets for two datasets.}
    \label{distribution}
\end{figure}

\textbf{Second}, during meta-testing, most graph FSL models generally train a classifier directly on support data consisting of limited labeled examples from novel classes and then apply the trained classifier to the query set to evaluate the model performance \cite{tan2022transductive, wang2022task, kim2023task}. One potential concern with this approach arises from the minimal size of the support data, which may not sufficiently represent the data distribution of the larger, unlabeled query set. This \textit{distribution shift} between the support and query sets poses a significant risk. To validate our hypothesis, we visualize the data distributions of the support and query sets for a sampled meta-testing task of two commonly used graph datasets, CoraFull \cite{bojchevski2017deep} and WikiCS \cite{mernyei2020wiki}, employing the kernel density estimation. The results are shown in Fig. \ref{distribution}. We observe a clear distribution shift between the support and query sets, which inevitably hamper the model performance.

\textbf{Finally}, these graph FSL models almost universally assume access to abundant labeled data from base classes during meta-training, enabling diverse task composition. However, in practice, this is infeasible due to the prohibitive time and even expertise domain knowledge required for data annotation. For example, in the protein interaction graph, predicting the properties of compounds requires expensive wet lab experiments \cite{hu2019strategies}. In gene regulatory networks, precise annotation of gene functions is challenging even for veteran researchers \cite{shu2021modeling}. These examples underscore the necessity of developing approaches that eliminate reliance on labeled data from base classes.

To address the above limitations, we propose a novel model for unsupervised graph few-shot learning named STAR. Specifically, \textit{during meta-training}, drawing inspiration from the success of graph contrastive learning (GCL), we adopt it to extract meaningful instance-level representations from graphs. Next, for traditional supervised graph FSL models, it is straightforward to construct set-input tasks and input them into neural set functions to extract discriminative set-level features. However, in our focused scenario, where the labels of all base data are invisible, a serious challenge is how to construct sets from unlabeled nodes. One natural solution is to apply multiple random data augmentations to the entire graph, similar to discrete images, and group nodes with same origins into sets. However, this approach requires storing all augmented graphs, incurring  extremely memory-cost, especially for large-scale graphs. Thus, to address this challenge, we reuse the two augmented views generated during the GCL process in the previous stage. For each node in one view, we retrieve the top-$k$ most similar nodes in the other view and split them into two sets. These two sets form positive pairs, while the remaining generated sets constitute negative pairs. We then perform set-level GCL on them to extract set-level features. \textit{During meta-testing}, our objective is to minimize the distribution shift between the support and query sets. This objective calibrates with the principles of optimal transport, which aims to derive the most efficient solution for transporting between two distributions. 
Therefore, 
we calibrate the distributions of the two sets based on the optimal transport and then use the calibrated support set to train the classifier, thereby mitigating the adverse effects of distribution shift. Empirically, even without introducing complicated techniques, our approach consistently outperforms other methods across multiple datasets. Theoretically, we prove that our approach can extract more task-relevant information and improve the model generalizability.
In summary, our main contributions are listed below.

(I) We propose a novel model for unsupervised graph few-shot learning, namely STAR. It can effectively obtain set-level features by set functions and mitigate the distribution shift by optimal transport.

(II) We theoretically prove STAR can capture more task-relevant information and tight the upper bound of generalization error.

(III) We empirically conduct extensive experiments on several benchmark datasets and achieve the state-of-the-art performance compared to other models.

%% file: related_work.tex
\section{Related Work}
\textbf{Graph Few-shot Learning.}
Graph FSL aims to enable models to quickly generalize previously learned meta-knowledge to novel tasks given limited training data, with wide applications in real-world scenarios \cite{zhang2022few, wang2023contrastive, tan2023virtual}. These models can mainly be divided into two categories: optimization-based \cite{huang2020graph, liu2022few, liu2024meta} and metric-based \cite{ding2020graph, liu2024simple} graph FSL models. The former primarily focuses on how to better integrate GNNs with classical meta-learning algorithms like MAML \cite{finn2017model}. The latter aims to learn an effective metric space for measuring the distance among massive unlabeled nodes and a few labeled nodes. Notably, most current developed graph FSL models extract transferable knowledge during the meta-training phase from diverse tasks composed of numerous labeled nodes from base classes. There are three representative ones, G-Meta \cite{huang2020graph}, Meta-GPS \cite{liu2022few}, and TEG \cite{kim2023task}. G-Meta constructs separate subgraphs for each node, where node-specific information is learned based on second-order optimization algorithms within these subgraphs. Meta-GPS learns the generalizable knowledge of base nodes by utilizing parameter initialization and task scaling and shifting transformations. TEG utilizes equivariant neural networks to capture the inductive bias of tasks, thereby reducing reliance on labeled nodes. Several models \cite{tan2022transductive, tan2023virtual} attempt to reduce the requirement for many labeled nodes from base classes. TLP \cite{tan2022transductive} directly applies the pretrained graph encoder to obtain node embeddings from novel classes for downstream tasks. However, the above methods merely extract instance-level features.

\noindent \textbf{Set Representation.} 
Considering that many machine learning tasks involve dealing with set-input problems, such as point cloud classification \cite{uy2019revisiting}, multi-instance learning \cite{ilse2018attention}, and sequence sorting \cite{vinyals2015order}, set neural network architectures \cite{zaheer2017deep, lee2019set} with permutation invariant properties have undergone significant development. DeepSets \cite{zaheer2017deep} processes elements and aggregates them using operations like max or sum pooling to obtain permutation-invariant set encodings. Set Transformer \cite{lee2019set} goes further by leveraging the expressive Transformer architecture to model pairwise interactions among elements in the set. Since FSL can also be viewed as set-level problems, several methods \cite{gordon2018meta, triantafillou2019meta} in the image processing have proposed enhancing few-shot model performance by employing sophisticated techniques to learn discriminative set representations. 
Notably, our model can be seamlessly adapted to any set encoder, showcasing its high adaptability.

\noindent \textbf{Optimal Transport.} 
The purpose of optimal transport is to find the minimum-cost solution for transporting two distributions \cite{torres2021survey}. Due to its effectiveness in comparing two probability distributions and generating optimal mappings, it has been widely used in fields such as computer vision \cite{arjovsky2017wasserstein, salimans2018improving, poulakakis2024beclr}, natural language processing \cite{yurochkin2019hierarchical, chen2019improving}, and domain adaptation \cite{courty2016optimal, courty2017joint}. While the methods based on optimal transport are competitive with state-of-the-art approaches in various fields, a notable drawback is the high computational expense involved in solving the optimal transport problem. 
A current popular approach is to introduce entropy regularization to approximate the optimal mapping function \cite{cuturi2013sinkhorn}. Recently, some works \cite{guo2021learning, guo2022adaptive} have proposed using optimal transport-based methods to transfer statistical information from base classes to calibrate the distribution of novel classes in FSL problems, achieving promising results. Notably, the key difference lies in the fact that we transport the distribution of support sets from the novel classes to the region of the query set distribution.

%% file: preliminary.tex
\section{Preliminary}
Given a graph $\mathcal{G}\!=\!\{\mathcal{V}, \mathcal{E}, \mathrm{X}, \mathrm{A}\}$, where $\mathcal{V}\!=\!\{v_1,\cdots,v_n\}$ and $\mathcal{E}\!=\!\{e_1,\cdots,e_m\}$ denote sets of nodes and edges, respectively. $\mathrm{X}\!\in\!\mathbb{R}^{n\times d}$ is the node feature matrix, and $\mathrm{A}\!\in\!\mathbb{R}^{n\times n}$ is the adjacency matrix. Here, we primarily consider the most popular \textit{few-shot node classification} in graph FSL to evaluate the model performance. In traditional \textit{supervised few-shot node classification}, the meta-training task set $\mathcal{D}_{tra}$ consist of many sampled tasks $\mathcal{T}_{tra}\!=\!\{\mathcal{S}_{tra},\mathcal{Q}_{tra}\}$, where $\mathcal{S}_{tra}\!=\!\{(v_i^{tra},y_i^{tra})\}_{i=1}^{NK}$ and $\mathcal{Q}_{tra}\!=\!\{v_i^{tra},y_i^{tra}\}_{i=1}^{NQ}$. Notably, $y_i^{tra}\!\in\!\mathcal{Y}_{tra}$, where $\mathcal{Y}_{tra}$ denotes the base classes. The meta-testing task $\mathcal{T}_{tes}\!=\!\{\mathcal{S}_{tes},\mathcal{Q}_{tes}\}\!=\!\{\{(v_i^{tes},y_i^{tes})\}_{i=1}^{NK}, \{v_i^{tes},y_i^{tes}\}_{i=1}^{NQ}\}$ share the same task construction as meta-training tasks, with the only difference being that the labels of nodes belong to novel classes $\mathcal{Y}_{tes}$ that are disjoint from $\mathcal{Y}_{tra}$, \textit{i.e.}, $\mathcal{Y}_{tra}\cap\mathcal{Y}_{tes}\!=\!\emptyset$. When the support set $\mathcal{S}_{tes}$ contains $N$ classes, and each class has $K$ nodes, it is referred to as an $N$-way $K$-shot task. Generally, graph FSL models are trained on $\mathcal{D}_{tra}$ and then fine-tuned on $\mathcal{S}_{tes}$, and its final performance is evaluated on $\mathcal{Q}_{tes}$. In our focused \textit{unsupervised few-shot node classification} scenarios, the meta-testing phase is the same as described above. The biggest difference lies in the meta-training phase, where we discard the node labels and train on a unified \textbf{unlabeled} data set $\mathcal{D}_{tra}\!=\!\{v_i\}_{i=1}^{sum}$. We summarize the main symbols used in Table \ref{notation}.

\begin{table}[ht]
    \caption{Detailed descriptions of used important symbols.}
    \label{notation}
    \centering
    \begin{tabular}{@{}c|c@{}}
    \toprule
       Symbols & Descriptions \\ \midrule
       $\mathcal{G}, \mathcal{V}, \mathcal{E}$  &  graph, node set, and edge set \\
       $\mathrm{X}, \mathrm{A}$ & node features, adjancency matrix \\
       $\mathrm{\hat{D}}, \mathrm{H}, \mathrm{S}$ & degree matrix, node and set embeddings \\
       $\mathrm{H}^\prime, \mathrm{S}^\prime$ & normalized node and set embeddings \\
       $\mathrm{\tilde{H}}, \mathrm{\tilde{S}}$ & refined node and set embeddings \\
       $\mathrm{Z}$ & final node embeddings \\
       $\mathrm{Z}_{spt}, \mathrm{Z}_{qry}$ & support and query embeddings \\
       $\mathrm{\hat{Z}}_{spt}$ & transported support embeddings \\
       $\Phi(\cdot)$ & neural set architecture \\
       $\phi(\cdot), \psi(\cdot)$ & two projectors \\
       $\mathcal{T}_{tes}, \mathcal{S}_{tes}, \mathcal{Q}_{tes}$ & meta-testing task, support set, query set \\
       $\mathbb H(\cdot), \lambda^\ast$ & entropy, optimal transport plan \\
       \bottomrule
    \end{tabular} %
\end{table}

%% file: method.tex
\section{Method}
\begin{figure*}
    \centering
    \includegraphics{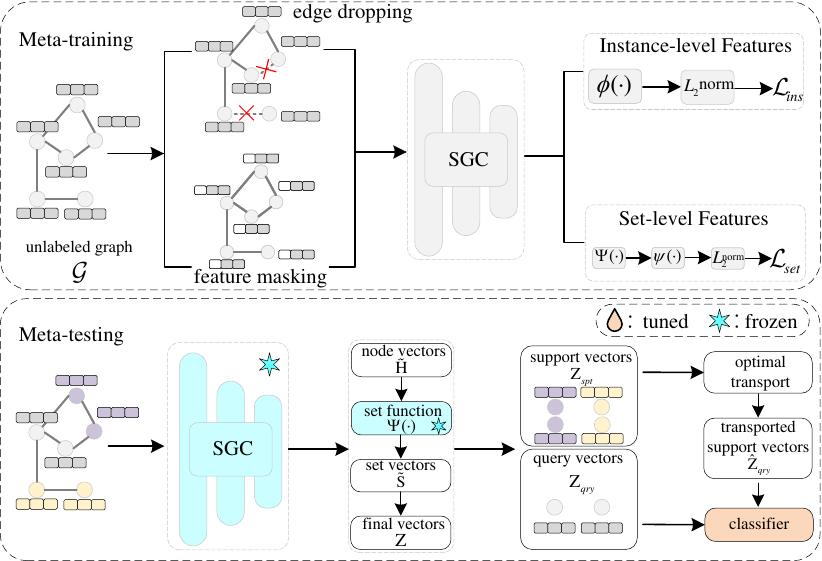}
    \caption{The overall framework of our model.}
    \label{framework}
\end{figure*}

In this section, we detail our proposed model, which consists of three parts: graph contrastive learning for instance-level features, graph contrastive learning for set-level features, and optimal transport for distribution calibration. To facilitate understanding, we present the framework of STAR in Fig. \ref{framework}.

\subsection{Graph Contrastive Learning for Instance-level Features}
We perform two stochastic data augmentation on the original graph $\mathcal{G}$, such as edge dropping or feature masking \cite{you2020graph}, resulting in two augmented graphs $\mathcal{G}_1$ and $\mathcal{G}_2$. Next, we input them into the graph encoder with shared weights to obtain node embeddings.
Following a simple principle, we choose SGC \cite{wu2019simplifying} as the graph encoder, which eliminates nonlinear activation functions between graph convolutional layers while collapsing weights into a linear transformation. The concrete procedure can be formulated as:
\begin{equation}
\label{sgc}
    \mathrm{H}=\tilde{\mathrm{A}}\cdots\tilde{\mathrm{A}}\mathrm{X}\mathrm{W}^{(0)}\cdots\mathrm{W}^{(\ell-1)}=\tilde{\mathrm{A}}^\ell\mathrm{X}\mathrm{W}^\ast,
\end{equation}
where $\tilde{\mathrm{A}}\!=\!\hat{\mathrm{D}}^{-1/2}\hat{\mathrm{A}}\hat{\mathrm{D}}^{-1/2}$ denotes the normalized adjacency matrix. $\hat{\mathrm{A}}\!=\!\mathrm{A}+\mathrm{I}$ and $\hat{\mathrm{D}}_i\!=\!\sum_j\hat{\mathrm{A}}_{ij}$ are the adjacency matrix with self-loops and the corresponding degree matrix. $\mathrm{W}^\ast\!\in\!\mathbb R^{d\times d^\prime}$ is the trainable parameters. $\mathrm{H}\!\in\!\mathbb R^{n\times d^\prime}$ is the learned node embeddings. 
Then, we utilize a projector $\phi(\cdot)$ to map node embeddings into a latent space where contrastive learning is applied, followed by $L_2$ normalization, \textit{i.e.}, $\mathrm{H}^\prime\!=\!\phi(\mathrm{H}), \mathrm{H}^\prime\!=\!\mathrm{H}^\prime/||\mathrm{H}^\prime||_2$.

We consider pairs of nodes originating from the same node as positive samples, and pairs from different nodes as negative samples. Thus, the instance-level graph contrastive loss for two graphs with $2n$ nodes is defined as:
\begin{equation}
\label{gcl_ins}
    \mathcal{L}_{ins}=-\frac{1}{2n}\sum_{i=1}^{2n}\log\frac{\exp((\mathrm{H}^\prime_i\cdot\mathrm{H}^\prime_j)/\tau)}{\sum_{k,k\neq i}\exp((\mathrm{H}^\prime_i\cdot\mathrm{H}^\prime_k)/\tau)},
\end{equation}
where $(\mathrm{H}^\prime_i, \mathrm{H}^\prime_j)$ and $(\mathrm{H}^\prime_i, \mathrm{H}^\prime_k)$ are defined as positive and negative sample pairs, respectively. $\tau$ is the temperature.
\subsection{Graph Contrastive Learning for Set-level Features}
In the previous stage, only instance-level features are considered, neglecting set-level features. Given that graph FSL problems are interpreted as set-level problems, we propose to construct two sets for each node from one view to form positive set pairs, where each set consists of nodes most similar to the target one. Specifically, for each node $v_i$ in $\mathcal{G}_1$, we retrieve the top-$k$ most similar nodes from $\mathcal{G}_2$ by the dot product (\textit{i.e.}, $\mathrm{H}_i\mathrm{H}^\top_{\mathcal{G}_2}$) to form a set $\{\mathrm{H}_{i,m}\}_{m=1}^k\!\in\!\mathbb R^{k\times d^\prime}$. The underlying rationale is that the most essential graph information should remain invariant across different perspectives. Then, we split the set into two equal parts $\Omega_i$ and $\Omega_j\!\in\!\mathbb R^{k/2\times d^\prime}$ 
and feed them into a neural set function $\Psi(\cdot)$ to derive the set representations. Here, we adopt a simple neural set architecture, which is defined as:
\begin{equation}
\label{deepset}
    \mathrm{S}_i=\Psi(\Omega_i)=\mathrm{MLP}(\text{sumpool}(\Omega_i)),
\end{equation}
where $\mathrm{S}_i\!\in\!\mathbb R^{d^\prime}$ is the obtained set representation, and  ``sumpool'' denotes the sum pooling operation.

Subsequently, we employ another projector $\psi(\cdot)$ to map set representations and perform $L_2$ normalization, \textit{i.e.}, $\mathrm{S}^\prime\!=\!\psi(\mathrm{S})$, $\mathrm{S}^\prime\!=\!\mathrm{S}^\prime/||\mathrm{S}^\prime||_2$. 

The set-level graph contrastive loss for $2n$ sets is formulated as: 
\begin{equation}
\label{set_level}
    \mathcal{L}_{set}=-\frac{1}{2n}\sum_{i=1}^{2n}\log\frac{\exp((\mathrm{S}^\prime_i\cdot\mathrm{S}^\prime_j)/\tau)}{\sum_{k,k\neq i}\exp((\mathrm{S}^\prime_i\cdot\mathrm{S}^\prime_k)/\tau)},
\end{equation}
where $(\mathrm{S}^\prime_i, \mathrm{S}^\prime_j)$ and $(\mathrm{S}^\prime_i, \mathrm{S}^\prime_k)$ are defined as positive and negative set pairs, respectively.

The used loss function in the meta-training stage is $\mathcal{L}\!=\!\mathcal{L}_{ins}\!+\!\mathcal{L}_{set}$. After conducting meta-training, we keep the pretrained graph encoder and neural set function while discarding other components. Next, we input the original graph $\mathcal{G}$ into the pre-trained graph encoder to obtain node embeddings $\tilde{\mathrm{H}}\!\in\!\mathbb R^{n\times d^\prime}$. We then perform a dot product operation on $\tilde{\mathrm{H}}$ to retrieve the top-$k$ similar nodes for each node in the graph. These retrieved nodes are subsequently fed into the pre-trained neural set function to obtain set representations
$\tilde{\mathrm{S}}\!\in\!\mathbb R^{n\times d^\prime}$. 
Finally, we merge $\tilde{\mathrm{H}}$ and $\tilde{\mathrm{S}}$ together to obtain updated node embeddings $\mathrm{Z}\!=\!\tilde{\mathrm{H}}||\tilde{\mathrm{S}}\!\in\!\mathbb R^{n\times 2d^\prime}$ for downstream analysis.
\subsection{Optimal Transport for Distribution Calibration}
Given a meta-test task $\mathcal{T}_{tes}\!=\!\{\mathcal{S}_{tes},\mathcal{Q}_{tes}\}$, we can extract the support embeddings $\mathrm{Z}_{spt}\!\in\!\mathbb R^{NK\times d^\prime}$ and query embeddings $\mathrm{Z}_{qry}\!\in\!\mathbb R^{NQ\times d^\prime}$. Considering the distribution shift between the support and query sets, directly using the acquired support embeddings to train the classifier and then use the classifier to evaluate the query set would yield suboptimal results. Hence, we utilize the philosophy of optimal transport to calibrate two distributions. Specifically, we adopt the obtained optimal transport solution $\lambda^\ast$ to map the support data $\mathcal{S}_{tes}$ into the distribution space of the query data $\mathcal{Q}_{tes}$. The distribution transition problem can be formulated as:
\begin{equation}
    \underset{{\lambda\in\Lambda}}\min\langle\lambda,\mathcal{D}\rangle=\underset{{\lambda\in\Lambda}}\min\sum\nolimits_{ij}\lambda_{ij}\mathcal{D}_{ij},
\end{equation}
where $\mathcal{D}$ denotes the pairwise distance matrix formed by the elements from $\mathrm{Z}_{spt}$ and $\mathrm{Z}_{qry}$. The set of transport solutions $\Lambda$ satisfies:
\begin{equation}
\label{op_condition}
    \Lambda = \{\lambda\in\mathbb R_+^{NK\times NQ}|\lambda\textbf{1}_{NQ}=\alpha, \lambda^\top\textbf{1}_{NK}=\beta\},
\end{equation}
where $\mathbf{1}_{NQ}$ is a vector of dimension $NQ$ with all elements being 1. $\alpha$ and $\beta$ are distributions of $\mathrm{Z}_{spt}$ and $\mathrm{Z}_{qry}$, respectively. 

To effectively solve the aforementioned optimization problem, we employ the Sinkhorn-Knopp algorithm \cite{cuturi2013sinkhorn}. The optimal transition solution can be defined as: 
\begin{equation}
\label{optimal_transport}
    \lambda^\ast=\underset{{\lambda\in\Lambda}}{\text{argmin}}\langle\lambda, D\rangle-\epsilon\mathbb H(\lambda),
\end{equation}
where $\mathbb H(\lambda)\!=\!-\sum_{ij}\lambda_{ij}\log\lambda_{ij}$ is the entropy function. Particularly, with $\alpha\!=\!\frac{1}{NK}\textbf{1}_{NK}$ and $\beta\!=\!\frac{1}{NQ}\textbf{1}_{NQ}$, this is to ensure that $\mathrm{Z}_{spt}$ is evenly partitioned into $NQ$ parts.

Next, we utilize $\lambda^\ast$ to map the support embeddings to the region of the query embeddings, \textit{i.e.}, 
$\hat{\mathrm{Z}}_{spt}=\lambda^{\ast\top}\mathrm{Z}_{spt}$,
where $\hat{\mathrm{Z}}_{spt}$ is the transported support embeddings.

Finally, we train a linear classifier on $\hat{\mathrm{Z}}_{spt}$ and use it to evaluate the model performance on $\hat{\mathrm{Z}}_{qry}$. Our training algorithm can be found 
in Algorithm \ref{pseudo_code}. Moreover, we also provide the complexity analysis of the proposed model in \textbf{Appendix} \ref{complexity}.

\begin{algorithm}
\caption{Training procedure of STAR}
\label{pseudo_code}
\renewcommand{\algorithmicrequire}{\textbf{Input:}}
\renewcommand{\algorithmicensure}{\textbf{Output:}}
\begin{algorithmic}[1]
\REQUIRE A graph $\mathcal{G}\!=\!\{\mathcal{V},\mathcal{E},\mathrm{Z},\mathrm{A}\}$.\\
\ENSURE The well-trained STAR.
\STATE // \textit{Meta-training process}
    \WHILE{\textit{not convergence}}
    \STATE Perform data augmentation to obtain $\mathcal{G}_1$ and $\mathcal{G}_2$.
    \STATE Learn node embeddings using Eq.\ref{sgc}.
    \STATE Utilize a projector $\phi(\cdot)$ followed by $L_2$ normalization.
    \STATE Perform instance-level contrastive loss $\mathcal{L}_{ins}$ with Eq.\ref{gcl_ins}.
    \STATE Retrieve top-$k$ similar nodes to form a set.
    \STATE Obtain the set embeddings using Eq.\ref{deepset}.
    \STATE Utilize a projector $\phi(\cdot)$ followed by $L_2$ normalization.
    \STATE Perform set-level contrastive loss $\mathcal{L}_{set}$ with Eq.\ref{set_level}.
    \STATE Optimize the model using $\mathcal{L}=\mathcal{L}_{ins}+\mathcal{L}_{set}$.
    \STATE Obtain the final node embeddings $\mathrm{Z}$.
    \ENDWHILE
    \STATE // \textit{Meta-testing process}
    \STATE Construct meta-testing task $\mathcal{T}_{tes}$.
    \STATE Compute the support and query embeddings $\mathrm{Z}_{spt}$ and $\mathrm{Z}_{qry}$.
    \STATE Conduct optimal transport to obtain the optimal plan $\lambda^\ast$.
    \STATE Obtain transported support embeddings $\hat{\mathrm{Z}}_{spt}$ using $\lambda^\ast$.
    \STATE Train the classifier with $\hat{\mathrm{Z}}_{spt}$
    \STATE Predict the node labels in $\mathcal{Q}_{tes}$ with the trained classifier.
\end{algorithmic}
\end{algorithm}

%% file: theory.tex
\section{Theoretical Analysis}
\label{theoretical_analysis}
In this section, we provide the theoretical foundation for our proposed model. We prove that STAR can capture more task-relevant information and has tighter upper bounds of generalization error. We provide the following theorems to support our argument.
\begin{theorem}
\label{mutual_information}
    For the classification task $\mathrm{T}$, let $\mathrm{Z}$ denotes the node representations obtained under our training objective, and $\tilde{\mathrm{H}}$ and $\tilde{\mathrm{S}}$ denote the node representations obtained individually under the instance-level and set-level loss, respectively. Then we have the following inequality holds:
    \begin{equation}
        \mathrm{I}(\mathrm{Z};\mathrm{T})\geq \max\{\mathrm{I}(\tilde{\mathrm{H}};\mathrm{T}), \mathrm{I}(\tilde{\mathrm{S}};\mathrm{T})\},
    \end{equation}
    where $\mathrm{I}(\cdot;\cdot)$ is the mutual information.
\end{theorem}
Theorem \ref{mutual_information} indicates that $\mathrm{Z}$ takes into account both instance-level and set-level features, thereby incorporating more task-relevant information compared to considering them individually. Naturally, we can readily obtain the following corollary. 
\begin{corollary}
\label{task_error}
    The classifier trained on $\mathrm{Z}$ exhibits smaller error upper bounds than that of trained independently on $\mathrm{\tilde{H}}$ and $\mathrm{\tilde{S}}$, defined as:
    \begin{equation}
        \mathrm{U}(\mathrm{P}(\mathrm{Z})) \leq \min\{\mathrm{U}(\mathrm{P}(\mathrm{\tilde{H}})), \mathrm{U}(\mathrm{P}(\mathrm{\tilde{S}}))\},
    \end{equation}
    where $\mathrm{U}(\cdot)$ denotes the upper bound of the function and $\mathrm{P}(\mathrm{Z})=\mathbb E_{\mathrm{Z}}(\min\mathrm{C}(y|\mathrm{Z}))$ is the Bayes error rate of the given representations, which measures the optimal performance of the learned classifier. $\mathrm{C}(y|\mathrm{Z})$ denotes conditional risk on the representation $\mathrm{Z}$.
\end{corollary}
Corollary \ref{task_error} further highlights the effectiveness of our proposed model, which can achieve smaller classification error and benefit the downstream task.

Before presenting the generalization error upper bound of STAR, we first provide some definitions.
Let the expected risk $\mathsf R$ and the corresponding empirical risk $\mathsf{\hat{R}}$ be denoted as,
\begin{equation}
\begin{aligned}
     &\mathsf{R}=\mathbb E_{\mathcal{T}_{tes}\sim p(\mathcal{T})}\mathbb E_{(\mathrm{Z}_{sup},\mathrm{Y}_{sup})\sim\mathcal{T}_{tes}}\mathcal{L}_{cr}(\theta^\top\mathrm{Z}_{sup},\mathrm{Y}_{sup}), \\
     &\mathsf{\hat{R}}=\mathbb E_{\mathcal{T}_{tes}\sim \hat{p}(\mathcal{T})}\mathbb E_{(\mathrm{Z}_{sup},\mathrm{Y}_{sup})\sim \hat{p}(\mathcal{T}_{tes})}\mathcal{L}_{cr}(\theta^\top\mathrm{Z}_{sup},\mathrm{Y}_{sup})\\
     &=\frac{1}{t}\sum_{i=1}^t\frac{1}{NK}\sum_{j=1}^{NK}\mathcal{L}_{cr}(\theta^\top\mathrm{Z}_{i,j},\mathrm{Y}_{i,j})
\end{aligned}
\end{equation}
where $\mathcal{L}_{cr}(\cdot)$ is the cross entropy loss and $\theta$ is the learnable matrix. 
Following previous work \cite{yao2021improving, yao2021meta}, the function class $\mathcal{F}_\gamma$ is given by $\mathcal{F}_\gamma\!=\!\{\mathrm{Z}\rightarrow \theta^\top\mathrm{Z}:\theta\Sigma\theta^\top\leq\gamma\}$, where $\Sigma\!=\!\mathbb E[\mathrm{Z}\mathrm{Z}^\top]$. Next, we formally give Theorem \ref{upper_bound} that specify the generalization gap as follows:
\begin{theorem}
\label{upper_bound}
    Suppose that $\mathrm{Z}$ and $\theta$ are bounded. Let $rank(\cdot)$ denote the rank of the matrix and 
    the samples be drawn from a distribution. For any $\delta\in(0,1)$, with the probability at least $1-\delta$ over the draw of sample, we have the following generalization bound holds,
    \begin{equation}
        |\mathsf{R}-\mathsf{\hat{R}}|\leq2\sqrt{\frac{\gamma\cdot \text{rank}(\Sigma)}{NK}}+\sqrt{\frac{\log(1/\delta)}{2NK}}.
    \end{equation}
\end{theorem}
Theorem \ref{upper_bound} suggests that the generalization gap is related to the weight constraint parameter $\gamma$. Therefore, we can introduce a regularization term into the objective function $\mathcal{L}_{cr}$ during the meta-testing phase to obtain a smaller $\gamma$, thereby reducing the generalization gap and resulting in better generalization. All detailed proof can be found in Appendix \ref{proof}.

%% file: experiment.tex
\section{Experiment}
\subsection{Dataset}
To empirically validate the effectiveness of our proposed model, we utilize several widely adopted datasets for few-shot node classification tasks, including \textbf{CoraFull} \cite{bojchevski2017deep}, \textbf{Coauthor-CS} \cite{shchur2018pitfalls}, \textbf{Cora} \cite{yang2016revisiting}, \textbf{WikiCS} \cite{mernyei2020wiki}, \textbf{ML} \cite{bojchevski2017deep}, and \textbf{CiteSeer} \cite{yang2016revisiting}. Moreover, we adopt two large-scale datasets, consisting of \textbf{ogbn-arxiv} \cite{hu2020open} and \textbf{ogbn-products} \cite{hu2020open}. The statistics of these datasets are presented in Table \ref{dataset}. We provide detailed descriptions of the evaluated datasets in \textbf{Appendix} \ref{dataset_detail}.

\begin{table}
\centering
\caption{Statistics of the datasets.}
\label{dataset}
\resizebox{0.4\textwidth}{!}{%
\begin{tabular}{@{}c|cccc@{}}
\toprule
Dataset     & \# Nodes & \# Edges & \# Features & \# Labels \\ \midrule
CoraFull   & 19,793   & 65,311   & 8,710       & 70        \\
Coauthor-CS & 18,333   & 81,894   & 6,805       & 15        \\
Cora        & 2,708    & 5,278    & 1,433       & 7         \\
WikiCS     & 11,701   & 216,123  & 11,701      & 10        \\
ML     & 2,995    & 16,316   & 2,879       & 7         \\
CiteSeer    & 3,327    & 4,552    & 3,703       & 6         \\ 
ogbn-arxiv & 169,343 & 1,166,243 & 128 & 40 \\
ogbn-products & 2,449,029 & 61,859,140 & 100 & 47 \\ \bottomrule
\end{tabular}%
}
\end{table}

\subsection{Baseline}
To demonstrate the superiority of our proposed model, we primarily select three types of baselines for comparison. \textit{Graph embedding methods} consist of \textbf{DeepWalk} \cite{perozzi2014deepwalk}, \textbf{node2vec} \cite{grover2016node2vec}, \textbf{GCN} \cite{kipf2016semi}, and \textbf{SGC} \cite{wu2019simplifying}. \textit{Traditional meta learning methods} contain \textbf{ProtoNet} \cite{snell2017prototypical} and \textbf{MAML} \cite{finn2017model}. \textit{Graph meta learning methods} include \textbf{Meta-GNN} \cite{zhou2019meta}, \textbf{GPN} \cite{ding2020graph}, \textbf{G-Meta} \cite{huang2020graph}, \textbf{TENT} \cite{wang2022task}, \textbf{Meta-GPS} \cite{liu2022few}, \textbf{X-FNC} \cite{wang2023few}, \textbf{TEG} \cite{kim2023task}, \textbf{COSMIC} \cite{wang2023contrastive}, and \textbf{TLP} \cite{tan2022transductive}. The descriptions of these baselines can be found in \textbf{Appendix} \ref{baseline_descrip}.

\subsection{Implementation Details}
In the \textit{GCL for instance-level features} stage, we employ 2-layer (\textit{i.e.}, $\ell=2$) SGC with 16 dimensions. We adopt edge dropping and feature masking techniques for augmentation, and the suitable ratios are determined by grid search from 0 to 0.4. In the \textit{GCL for set-level features} stage, the value of $k$ in the top-$k$ is specified as 20. All the projectors, including $\phi(\cdot)$ and $\psi(\cdot)$, are implemented by MLPs with one hidden layer of 16 dimensions followed by the ReLU activation function. The temperature parameter $\tau$ of all GCL is set to $0.5$. We use the Adam optimizer \cite{kingma2014adam} and the learning rate is 0.001. To prevent model overfitting, we adopt the early-stopping training strategy. 
To ensure fair comparison, we evaluate the model performance using the mean accuracy across 50 randomly sampled meta-testing tasks. We conduct each experiment five times and report the mean accuracy along with the standard deviations. For graph embedding methods, we train a linear classier after obtaining the node embeddings. For traditional meta learning and graph meta learning methods, we use the hyperparameters of their original work. All the experiments are conducted by NVIDIA 3090Ti GPUs. 

%% file: result.tex
\section{Results}
\subsection{Model Performance}
We present the experimental results of our proposed model and other baselines under various few-shot settings across several benchmarks in Tables \ref{res_many}, \ref{res_few}, and \ref{res_large}. Here, we conduct four few-shot setups for the first and last two datasets with more categories, and two few-shot setups for the middle four datasets with fewer categories. 
According to the above results, we can observe that STAR achieves outstanding performance on almost all datasets, including both small and large datasets, compared to other models, providing strong evidence of its effectiveness and robustness in this task. A plausible reason is that during the meta-training phase, we not only utilize instance-level features but also exploit set-level features inherent within the graph, effectively capturing the permutation-invariant features in set-input tasks. Additionally, during the meta-testing phase, we employ the optimal transport concept to calibrate the biased sampling of the support set distribution to match the query set distribution. The adopted strategies ensure our model's superiority over others.

Also, we find that graph meta learning models considerably outperform other types of models, which aligns with our expectations. 
A key advantage lies in their ability to effectively capture both the features and structural information inherent in graphs.
Moreover, their sophisticated algorithm designs are specifically tailored for addressing graph FSL problems, enabling them to rapidly adapt to downstream tasks with minimal labeled nodes. On the contrary, traditional meta-learning methods such as ProtoNet and MAML achieve unsatisfactory performance on all evaluated datasets. This is because these models completely neglect the crucial structural information within graphs, leading to significant decreases in their ability to generalize to graph-structural data.

\begin{table*}[ht]
\centering
\caption{Accuracies (\%) of different models in the first two datasets. Underline: runner-up.}
\begin{tabular}{@{}c|cccc|cccc@{}}
\toprule
\multirow{2}{*}{Model} &
  \multicolumn{4}{c|}{CoraFull} &
  \multicolumn{4}{c}{Coauthor-CS} \\ \cmidrule(l){2-9} 
 &
  5 way 3 shot &
  5 way 5 shot &
  10 way 3 shot &
  10 way 5 shot &
  2 way 3 shot &
  2 way 5 shot &
  5 way 3 shot &
  5 way 5 shot \\ \midrule
DeepWalk &
  23.62$\pm$3.99 &
  25.93$\pm$3.45 &
  15.32$\pm$4.12 &
  17.03$\pm$3.73 &
  59.52$\pm$2.72 &
  63.12$\pm$3.12 &
  33.76$\pm$3.21 &
  40.15$\pm$2.96 \\
node2vec &
  23.75$\pm$2.93 &
  25.42$\pm$3.61 &
  13.90$\pm$3.32 &
  15.21$\pm$2.64 &
  56.16$\pm$4.19 &
  60.22$\pm$4.06 &
  30.35$\pm$3.93 &
  39.16$\pm$3.79 \\
GCN &
  34.65$\pm$2.76 &
  39.83$\pm$2.49 &
  29.23$\pm$3.25 &
  34.14$\pm$2.15 &
  73.52$\pm$1.97 &
  77.20$\pm$3.01 &
  52.19$\pm$2.31 &
  56.35$\pm$2.99 \\
SGC &
  39.56$\pm$3.52 &
  44.53$\pm$2.92 &
  35.12$\pm$2.71 &
  39.53$\pm$3.32 &
  75.49$\pm$2.15 &
  79.63$\pm$2.01 &
  56.39$\pm$2.26 &
  59.25$\pm$2.16 \\ \midrule
ProtoNet &
  33.67$\pm$2.51 &
  36.53$\pm$3.76 &
  24.90$\pm$2.03 &
  27.24$\pm$2.67 &
  71.18$\pm$3.82 &
  75.51$\pm$3.19 &
  47.71$\pm$3.92 &
  51.66$\pm$2.51 \\
MAML &
  37.12$\pm$3.16 &
  47.51$\pm$3.09 &
  26.61$\pm$2.19 &
  31.60$\pm$2.91 &
  62.32$\pm$4.60 &
  65.20$\pm$4.20 &
  36.99$\pm$4.32 &
  42.12$\pm$2.43 \\ \midrule
Meta-GNN &
  52.23$\pm$2.41 &
  59.12$\pm$2.36 &
  47.21$\pm$3.06 &
  53.32$\pm$3.15 &
  85.76$\pm$2.74 &
  87.86$\pm$4.79 &
  75.87$\pm$3.88 &
  68.59$\pm$2.59 \\
GPN &
  53.24$\pm$2.33 &
  60.31$\pm$2.19 &
  50.93$\pm$2.30 &
  56.21$\pm$2.09 &
  85.60$\pm$2.15 &
  88.70$\pm$2.21 &
  75.88$\pm$2.75 &
  81.79$\pm$3.18 \\
G-Meta &
  57.52$\pm$3.91 &
  62.43$\pm$3.11 &
  53.92$\pm$2.91 &
  58.10$\pm$3.02 &
  92.14$\pm$3.90 &
  93.90$\pm$3.18 &
  75.72$\pm$3.59 &
  74.18$\pm$3.29 \\
TENT &
  64.80$\pm$4.10 &
  69.24$\pm$4.49 &
  51.73$\pm$4.34 &
  56.00$\pm$3.53 &
  89.35$\pm$4.49 &
  90.90$\pm$4.24 &
  78.38$\pm$5.21 &
  78.56$\pm$4.42 \\
Meta-GPS &
  65.19$\pm$2.35 &
  69.25$\pm$2.52 &
  61.23$\pm$3.11 &
  64.22$\pm$2.66 &
  90.16$\pm$2.72 &
  92.39$\pm$1.66 &
  81.39$\pm$2.35 &
  83.66$\pm$1.79 \\
X-FNC &
  69.32$\pm$3.10 &
  71.26$\pm$4.19 &
  49.63$\pm$4.45 &
  53.00$\pm$3.93 &
  90.95$\pm$4.29 &
  92.03$\pm$4.14 &
  \underline{82.93$\pm$2.02} &
  84.36$\pm$3.49 \\
TEG &
  72.14$\pm$1.06 &
  76.20$\pm$1.39 &
  61.03$\pm$1.13 &
  65.56$\pm$1.03 &
  \underline{92.36$\pm$1.59} &
  93.02$\pm$1.24 &
  80.78$\pm$1.40 &
  84.70$\pm$1.42 \\
COSMIC &
  \underline{73.03$\pm$1.78} &
  \underline{77.24$\pm$1.52} &
  \underline{65.79$\pm$1.36} &
  \underline{70.06$\pm$1.93} &
  89.35$\pm$4.49 &
  \underline{93.32$\pm$1.93} &
  78.38$\pm$5.21 &
  \underline{85.47$\pm$1.42} \\
TLP &
  66.32$\pm$2.10 &
  71.36$\pm$4.49 &
  51.73$\pm$4.34 &
  56.00$\pm$3.53 &
  90.35$\pm$4.49 &
  90.90$\pm$4.24 &
  82.30$\pm$2.05 &
  78.56$\pm$4.42 \\ \midrule
STAR &
  \textbf{77.77$\pm$0.10} &
  \textbf{81.24$\pm$0.98} &
  \textbf{68.60$\pm$0.63} &
  \textbf{73.53$\pm$0.49} &
  \textbf{95.20$\pm$1.36} &
  \textbf{96.43$\pm$0.47} &
  \textbf{88.08$\pm$0.26} &
  \textbf{90.59$\pm$0.35} \\ \bottomrule
\end{tabular}%
\label{res_many}
\end{table*}

\begin{table*}[ht]
\centering
\caption{Accuracies (\%) of different models in the middle four datasets. Underline: runner-up.}
\begin{tabular}{@{}c|cc|cc|cc|cc@{}}
\toprule
\multirow{2}{*}{Model} &
  \multicolumn{2}{c|}{Cora} &
  \multicolumn{2}{c|}{WikiCS} &
  \multicolumn{2}{c|}{Cora-ML} &
  \multicolumn{2}{c}{CiteSeer}\\ \cmidrule(l){2-9} 
 &
  \multicolumn{1}{c}{2 way 3 shot} &
  \multicolumn{1}{c|}{2 way 5 shot} &
  \multicolumn{1}{c}{2 way 3 shot} &
  \multicolumn{1}{c|}{2 way 5 shot} &
  \multicolumn{1}{c}{2 way 3 shot} &
  \multicolumn{1}{c|}{2 way 5 shot} &
  \multicolumn{1}{c}{2 way 3 shot} &
  \multicolumn{1}{c}{2 way 5 shot} \\ \midrule
DeepWalk & 36.70$\pm$2.99 & 41.51$\pm$2.70 & 37.29$\pm$2.51 & 38.59$\pm$2.49 & 40.32$\pm$3.10 & 42.19$\pm$2.55 & 39.72$\pm$3.42 & 43.22$\pm$3.19 \\
node2vec & 35.66$\pm$2.79 & 40.69$\pm$2.90 & 41.52$\pm$2.49 & 43.77$\pm$2.61 & 43.19$\pm$2.17 & 45.16$\pm$2.62 & 42.39$\pm$2.79 & 47.20$\pm$2.92 \\
GCN      & 69.96$\pm$2.52 & 67.95$\pm$2.36 & 72.55$\pm$2.31 & 74.55$\pm$2.91 & 67.45$\pm$2.11 & 69.22$\pm$2.01 & 53.79$\pm$2.39 & 55.76$\pm$2.56\\
SGC      & 70.15$\pm$1.99 & 70.67$\pm$2.11 & 74.15$\pm$2.57 & 76.34$\pm$2.59 & 69.51$\pm$3.10 & 70.29$\pm$1.95 &55.12$\pm$2.59 &57.25$\pm$2.79\\ \midrule
ProtoNet & 52.67$\pm$2.28 & 57.92$\pm$2.34 & 53.56$\pm$2.31 & 55.25$\pm$2.31 & 68.40$\pm$2.60 & 76.94$\pm$2.39 & 52.19$\pm$2.96 & 53.75$\pm$2.49\\
MAML     & 55.07$\pm$2.36 & 57.39$\pm$2.23 & 52.51$\pm$2.35 & 54.33$\pm$2.19 & 68.61$\pm$2.51 & 71.27$\pm$2.62 & 52.75$\pm$2.75 & 54.36$\pm$2.39\\ \midrule
Meta-GNN & 70.40$\pm$2.64 & 72.51$\pm$1.91 & 78.14$\pm$2.28 & 78.35$\pm$2.60 & 70.21$\pm$2.71 & 74.34$\pm$2.41 & 59.71$\pm$2.79 & 61.32$\pm$3.22\\
GPN      & 74.05$\pm$1.96 & 76.39$\pm$2.33 & 86.55$\pm$2.16 & \underline{87.80$\pm$1.95} & 80.70$\pm$2.41 & 83.21$\pm$2.15 &64.22$\pm$2.92 &65.59$\pm$2.49\\
G-Meta   & 74.39$\pm$2.69 & 80.05$\pm$1.98 & 61.09$\pm$2.84 & 78.35$\pm$2.60 & \underline{88.68$\pm$1.68} & \underline{92.16$\pm$1.14} & 57.59$\pm$2.42 & 62.49$\pm$2.30\\
TENT     & 58.25$\pm$2.23 & 66.75$\pm$2.19 & 68.85$\pm$2.42 & 70.35$\pm$2.26 & 55.65$\pm$2.19 & 58.30$\pm$2.05 & 65.20$\pm$3.19 & 67.59$\pm$2.95\\
Meta-GPS & 80.29$\pm$2.15 & 83.79$\pm$2.10 & 85.72$\pm$2.10 & 87.05$\pm$1.35 & 87.91$\pm$2.12 & 91.66$\pm$2.52 & 69.95$\pm$2.02 &72.56$\pm$2.06\\
X-FNC     & 78.19$\pm$3.25 & 82.70$\pm$3.19 & 83.80$\pm$3.42 & 86.30$\pm$3.20 & 85.62$\pm$3.12 & 90.36$\pm$2.99 & 67.96$\pm$3.10 & 70.29$\pm$3.05\\
TEG     & \underline{80.65$\pm$1.53} & 84.50$\pm$2.01 & \textbf{86.59$\pm$2.32} & 87.70$\pm$2.49 & 67.90$\pm$2.10 & 71.10$\pm$2.02 & \underline{73.79$\pm$1.59} & \underline{76.79$\pm$2.12}\\
COSMIC     & 65.37$\pm$2.49 & 69.10$\pm$2.30 & 85.51$\pm$2.30 & 86.72$\pm$1.90 & 66.02$\pm$2.29 & 72.10$\pm$2.25 & 70.22$\pm$2.56 & 75.10$\pm$2.30\\
TLP     & 71.10$\pm$1.66 & \underline{86.15$\pm$2.19} & 83.09$\pm$2.72 & 70.35$\pm$2.26 & 85.32$\pm$2.12 & 58.30$\pm$2.05 & 71.10$\pm$2.17 & 75.55$\pm$2.03 \\ \midrule
STAR &
  \textbf{86.37$\pm$2.13} &
  \textbf{88.87$\pm$1.68} &
  \underline{86.40$\pm$1.92} &
  \textbf{87.93$\pm$0.68} &
  \textbf{91.20$\pm$1.12} &
  \textbf{95.83$\pm$0.97} &\textbf{76.50$\pm$2.12} &\textbf{79.43$\pm$0.45} \\ \bottomrule
\end{tabular}%
\label{res_few}
\end{table*}

\begin{table*}[ht]
\centering
\caption{Accuracies (\%) of different models in the last two large scale datasets. Underline: runner-up. OOM: out-of-memory.}
\label{res_large}
\begin{tabular}{@{}c|cccc|cccc@{}}
\toprule
\multirow{2}{*}{Model} & \multicolumn{4}{c|}{ogbn-arxiv}                                                   & \multicolumn{4}{c}{ogbn-products}                                                     \\ \cmidrule(l){2-9} 
                       & 5 way 3 shot            & 5 way 5 shot            & 10 way 3 shot & 10 way 5 shot & 5 way 3 shot        & 5 way 5 shot        & 10 way 3 shot       & 10 way 5 shot       \\ \midrule
DeepWalk               & 24.12$\pm$3.16              & 26.16$\pm$2.95              & 20.19$\pm$2.35    & 23.76$\pm$3.02    & 24.92$\pm$2.36          & 27.90$\pm$2.56          & 16.72$\pm$3.16          & 19.26$\pm$2.96          \\
node2vec               & 25.29$\pm$2.96              & 27.39$\pm$2.56              & 22.99$\pm$3.15    & 25.95$\pm$3.12    & 25.29$\pm$2.96          & 30.52$\pm$2.79          & 17.56$\pm$2.76          & 21.19$\pm$2.66          \\
GCN                    & 32.26$\pm$2.11              & 36.29$\pm$2.39              & 30.21$\pm$1.95    & 33.96$\pm$1.59    & 34.09$\pm$2.59          & 39.42$\pm$2.19          & 25.59$\pm$2.92          & 30.19$\pm$2.32          \\
SGC                    & 35.19$\pm$2.76              & 39.76$\pm$2.95              & 31.99$\pm$2.12    & 35.22$\pm$2.52    & 37.52$\pm$2.12          & 41.22$\pm$2.66          & 29.36$\pm$2.29          & 33.66$\pm$2.90          \\ \midrule
ProtoNet               & 39.99$\pm$3.28          & 47.31$\pm$1.71          & 32.79$\pm$2.22    & 37.19$\pm$1.92    & 29.12$\pm$1.92          & 30.52$\pm$2.02          & 21.59$\pm$2.59          & 27.62$\pm$2.26          \\
MAML                   & 28.35$\pm$1.49          & 29.09$\pm$1.62          & 30.19$\pm$2.97    & 36.19$\pm$2.29    & 35.09$\pm$1.66          & 40.16$\pm$2.15          & 26.19$\pm$2.36          & 29.39$\pm$2.66          \\ \midrule
Meta-GNN               & 40.14$\pm$1.94          & 45.52$\pm$1.71          & 35.19$\pm$1.72    & 39.02$\pm$1.99    & 49.19$\pm$1.62          & 52.31$\pm$2.05          & 34.72$\pm$2.79          & 40.12$\pm$2.95          \\
GPN                    & 42.81$\pm$2.34          & 50.50$\pm$2.13          & 37.36$\pm$1.99    & 42.16$\pm$2.19    & 55.62$\pm$2.39          & 57.20$\pm$2.56          & 39.63$\pm$1.93          & 42.49$\pm$2.19          \\
G-Meta                 & 40.48$\pm$1.70          & 47.16$\pm$1.73          & 35.49$\pm$2.12    & 40.95$\pm$2.70    & 60.53$\pm$3.26          & 62.42$\pm$2.90          & 46.12$\pm$1.79          & 49.09$\pm$2.15          \\
TENT                   & 50.26$\pm$1.73          & 61.38$\pm$1.72          & 42.19$\pm$1.16    & 46.29$\pm$1.29    & OOM                 & OOM                 & OOM                 & OOM                 \\
Meta-GPS               & 52.16$\pm$2.01          & 62.55$\pm$1.95          & 42.96$\pm$2.02    & 46.86$\pm$2.10    & {\ul 65.42$\pm$2.12}          & {\ul 66.52$\pm$2.19}          & {\ul 50.92$\pm$1.66}          & {\ul 53.12$\pm$1.62}          \\
X-FNC                  & 52.36$\pm$2.75          & 63.19$\pm$2.22          & 41.92$\pm$2.72    & 46.10$\pm$2.16    & 65.15$\pm$3.19          & 65.72 3.96          & 50.12 2.16          & 52.95 1.75          \\
TEG                    & {\ul 57.35$\pm$1.14}          & 62.07$\pm$1.72          & {\ul 47.41$\pm$0.63}    & {\ul 51.11$\pm$0.73}    & OOM                 & OOM                 & OOM                 & OOM                 \\
COSMIC                 & 52.98$\pm$2.19          & {\ul 65.42$\pm$1.69}          & 43.19$\pm$2.72    & 47.59$\pm$2.19    & OOM                 & OOM                 & OOM                 & OOM                 \\
TLP                    & 41.96$\pm$2.29          & 52.99$\pm$2.05          & 39.42$\pm$2.15    & 42.62$\pm$2.09    & OOM                 & OOM                 & OOM                 & OOM                 \\ \midrule
\textbf{STAR}          & \textbf{61.15$\pm$0.62} & \textbf{68.39$\pm$0.95} & \textbf{50.89$\pm$2.02}    & \textbf{55.80$\pm$2.01}    & \textbf{69.96$\pm$1.26} & \textbf{72.29$\pm$1.62} & \textbf{58.82$\pm$2.21} & \textbf{60.84$\pm$1.76} \\ \bottomrule
\end{tabular}
\end{table*}

\begin{table*}[ht]
\caption{Results of different model variants on all datasets. OOM: out-of-memory.}
\label{ablation}
\centering
\begin{tabular}{@{}c|cccccccc@{}}
\toprule
\multirow{2}{*}{Model} & CoraFull            & Coauthor-CS         & Cora                & WikiCS              & ML                  & CiteSeer   & ogbn-arxiv  & ogbn-products         \\ \cmidrule(l){2-9} 
                       & 5 way 5 shot        & 2 way 5 shot        & 2 way 5 shot        & 2 way 5 shot        & 2 way 5 shot        & 2 way 5 shot  & 5 way 3 shot  & 5 way 3 shot      \\ \midrule
\textit{w/o instance}  & 75.35$\pm$2.49          & 93.73$\pm$0.34          & 85.07$\pm$2.59          & 85.87$\pm$1.28          & 92.10$\pm$2.50          & 74.03$\pm$2.66     & 57.29$\pm$1.25 & 65.39$\pm$1.02     \\
\textit{w/o set}       & 79.17$\pm$0.73          & 95.43$\pm$0.46          & 86.95$\pm$1.16          & 86.40$\pm$0.96          & 94.76$\pm$0.17          & 77.40$\pm$1.82   & 60.19$\pm$0.79      & 67.29$\pm$1.36  \\
\textit{w/o op}        & 79.72$\pm$1.53          & 95.56$\pm$0.12          & 87.17$\pm$2.39          & 87.02$\pm$1.11          & 95.26$\pm$2.06          & 78.50$\pm$1.47     & 60.39$\pm$1.72   & 68.12$\pm$1.12   \\
\textit{w original}    & 80.65$\pm$0.75          & 95.03$\pm$0.41          & 86.20$\pm$0.92          & 86.71$\pm$0.76          & 95.02$\pm$0.90          & 79.09$\pm$0.40      & 60.28$\pm$0.92    & 69.02$\pm$1.26 \\
\textit{w aug}    & OOM          & OOM          & 87.20$\pm$0.99          & 86.79$\pm$0.79          & 95.32$\pm$0.92          & 79.19$\pm$0.46      & OOM    & OOM \\
Ours                   & \textbf{81.24$\pm$0.98} & \textbf{96.43$\pm$0.47} & \textbf{88.87$\pm$1.68} & \textbf{87.93$\pm$0.68} & \textbf{95.83$\pm$0.97} & \textbf{79.43$\pm$0.45}  & \textbf{61.15$\pm$0.62} & \textbf{69.96$\pm$1.26} \\ \bottomrule
\end{tabular}%
\end{table*}

\subsection{Ablation Study}
To validate the effectiveness of the strategies employed, we design several model variants under different few-shot scenarios across all datasets on different few-shot settings. (I) \textit{w/o instance}: We eliminate instance-level GCL and kept only set-level GCL. (II) \textit{w/o set}: We discard set-level GCL and retain the other one. (III) \textit{w/o op}: We omit the optimal transport strategy and train the classifier directly using the support set. (IV) \textit{w original}: We retrieve the top-$k$ similar nodes in the original graph. (v) \textit{w aug}: We performed $k$ rounds of augmentation on the original graph, grouping nodes with the same source into a set. We present the experimental results in Table \ref{ablation}.

Based on the results, we have the in-depth analysis below. \textit{First}, regardless of which part is eliminated, the model performance exhibits varying degrees of decline, indicating the importance of each designed module. \textit{Second}, meaningful instance-level features can help distinguish subtle differences between categories, thus playing a crucial role in graph FSL problems. However, set-level features are also crucial as they impart the necessary permutation invariance for the focused problem. \textit{Third}, the used optimal transport strategy can enhance model performance by alleviating data distribution shift between support and query sets. \textit{Fourth}, when retrieving the top-$k$ similar nodes to a target node within the original graph instead of the generated augmented views, the homophily assumption of graphs may impact the selection of similar nodes within the target node's 1-hop neighborhood. \textit{Finally}, when we explicitly perform data augmentation to construct sets for set-level GCL, we encounter memory constraints in moderately sized datasets. Furthermore, the performance on other datasets is even inferior to that of our model. This underscores the effectiveness of the retrieval strategy we adopted.

\begin{figure}
    \centering
    \subfigure[]{\label{distribution_op}
    \includegraphics[width=0.23\textwidth]{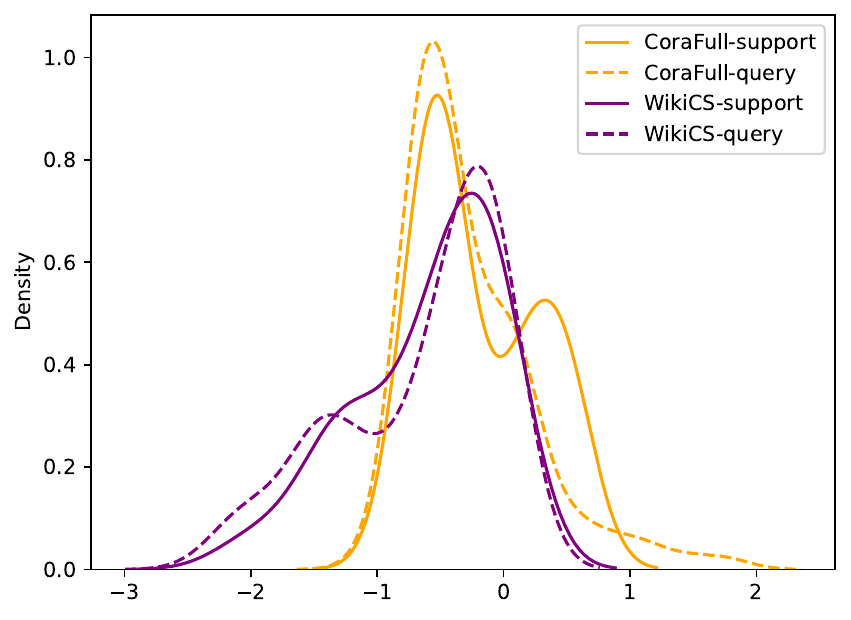}}
    \subfigure[]{\label{hyper}
    \includegraphics[width=0.23\textwidth]{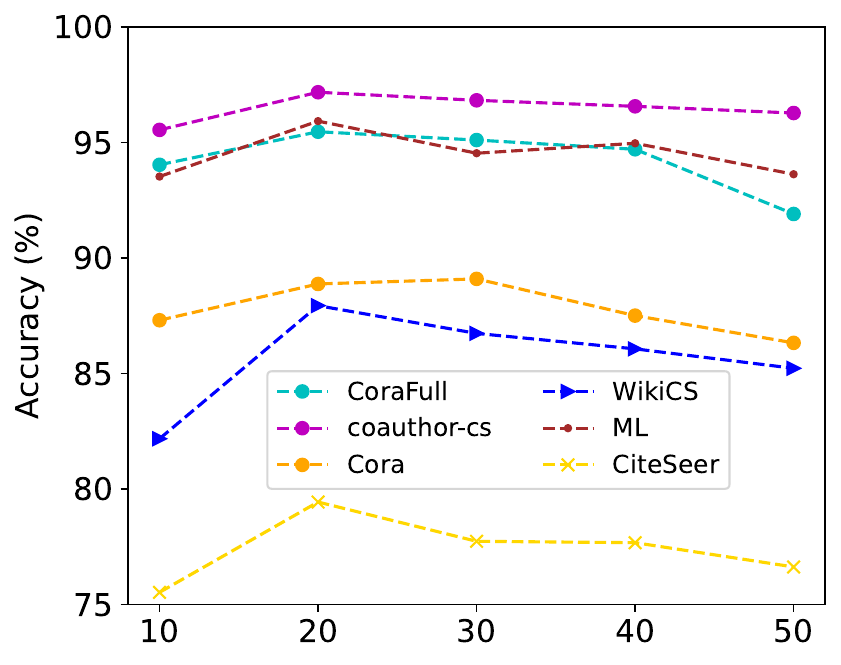}}
    \caption{(a): Distribution of support and query set after performing optimal transport. (b): Model performance varies with the value of $k$ in top-$k$ across all datasets.}
\end{figure}

Further, to provide a clearer depiction of the efficacy of optimal transport, we visualize the data distributions of support and query sets across various datasets after performing optimal transport strategies during the meta-testing phase, as shown in Fig. \ref{distribution_op}. We can observe that the distribution shift between the support and query sets has been significantly mitigated.

\subsection{Hyparameter Sensitivity}
We primarily investigate the impact of different values of $k$ in the top-$k$ retrieval on model performance under the 2-way 5-shot experimental setting. According to Fig. \ref{hyper}, we can find that the model performance exhibits a trend of initially increasing and then decreasing across all datasets, which is consistent with our expectations. One plausible reason is that when $k$ is too small, it fails to capture the inherent set features adequately, while when $k$ is too large, it introduces noise nodes dissimilar to the target node, thereby affecting the model performance.

\begin{figure}[h]
    \centering
    \subfigure[CiteSeer]{
    \includegraphics[width=0.23\textwidth]{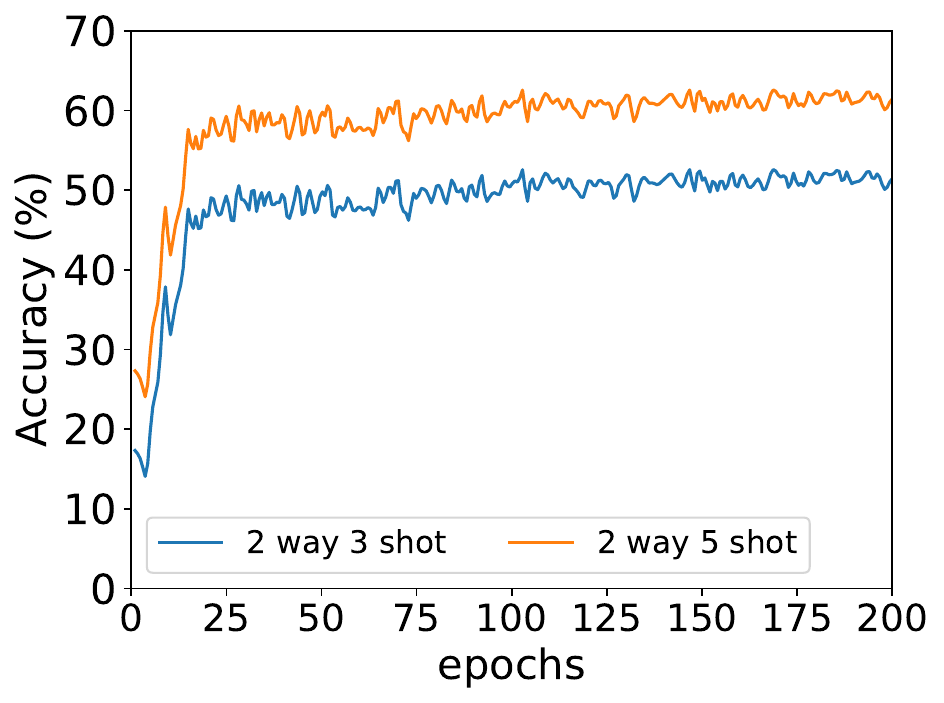}}
    \subfigure[ML]{
    \includegraphics[width=0.23\textwidth]{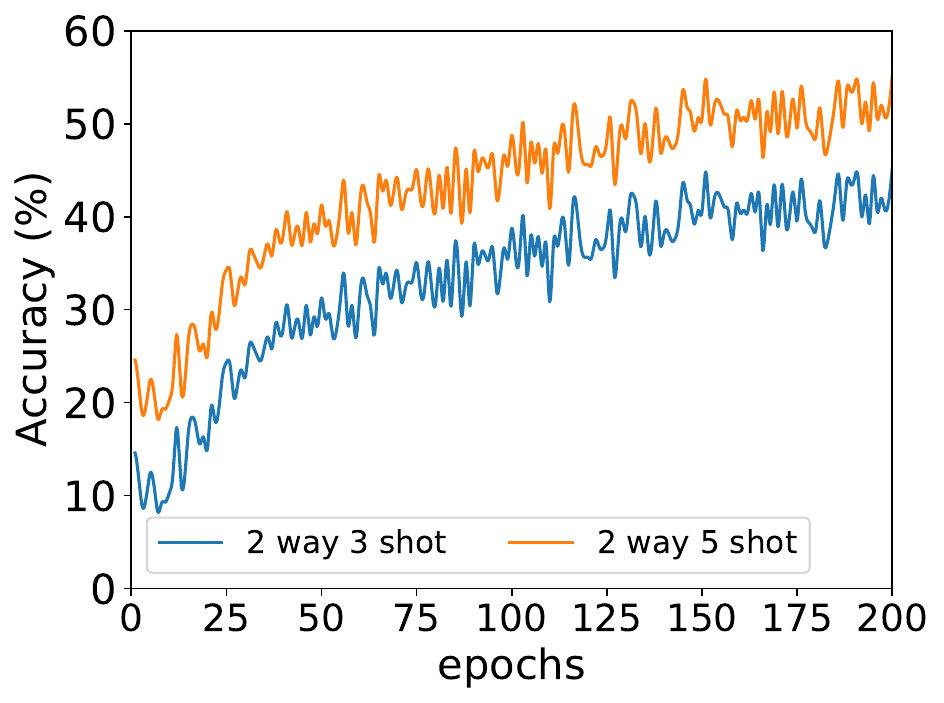}}
    \caption{Model performance varies with epochs across two datasets.}
    \label{trend}
\end{figure}

We also evaluate the accuracy of retrieved similar nodes for target nodes. We define a successful retrieval when the retrieved similar node shares a common label with the target node. We illustrate the trend of accuracy as epochs vary for the CiteSeer and ML datasets as shown in Fig. \ref{trend}. We observe that as the model training progresses, more similar nodes with labels consistent with the target node are retrieved. This indicates that this strategy indeed effectively captures semantic relationships among nodes.

%% file: conclusion.tex
\section{Conclusion}
In this work, we propose a novel model for unsupervised graph few-shot learning, namely STAR. In the meta-training stage, we first adopt GCL for extracting expressive instance-level features. Then, we capture the implicit set-level features inherent in the graph ignored by previous models. In the meta-testing stage, we employ the optimal transport strategy to calibrate the data distribution of support and query sets. Theoretically, we demonstrate that the derived node embeddings carry more task-relevant information, while providing an upper bound on the model generalizability. Empirically, we conduct sufficient experiments across various datasets on different few-shot settings, and the results show the superiority of our STAR compared to other models.

%% file: appendix.tex
\section{Supplementary Material}

\subsection{Complexity Analysis}
\label{complexity}
In this section, we conduct a detailed analysis of the time complexity of the proposed model. In the part of the graph contrastive learning for instance-level features, the time-consuming operations originate from SGC and instance-level graph contrastive learning, with the time complexity $O(\ell m^2+md+ndd^\prime)$ and $O(2n(d^\prime+(2n-2)d^\prime))$. Here, $\ell$, $n$, and $m$ denote the number of graph layers, nodes and edges. $d$ and $d^\prime$ denote the dimension of original and hidden features. In the part of graph contrastive learning for set-level features, the resulting time complexity mainly comprises neural set functions and set-level graph contrastive learning, which are $O(nd^\prime k+nd^{\prime2})$ and $O(2n(d^\prime+(2n-2)d^\prime))$ respectively. In the part of optimal transport for distribution calibration, the required time complexity is $O(\frac{NK\times NQ}{\epsilon^2})$, where $NK$ and $NQ$ denote the size of support and query embeddings. $\epsilon$ is the desired error for the computational cost. Based on the analysis above, the resulting time complexity is acceptable to us.

\subsection{Theoretical Proof}
\label{proof}
\subsubsection{Proof of Theorem \ref{mutual_information}}
Before presenting the formal proof, we establish that $\mathrm{I}(\cdot;\cdot)$ denotes the mutual information, and $\mathbb{H}(\cdot)$ denotes the entropy.
\begin{proof}
    To prove $\mathrm{I}(\mathrm{Z};\mathrm{T})\geq \max\{\mathrm{I}(\mathrm{\tilde{H}};\mathrm{T}),\mathrm{I}(\tilde{\mathrm S};\mathrm{T})\}$, we only need to demonstrate that $\mathrm{I}(\mathrm{Z};\mathrm{T})$ satisfies both $\mathrm{I}(\mathrm{Z};\mathrm{T})\geq \mathrm{I}(\mathrm{\tilde{H}};\mathrm{T})$ and $\mathrm{I}(\mathrm{Z};\mathrm{T})\geq\mathrm{I}(\tilde{\mathrm S};\mathrm{T})$ simultaneously.
    According to the definition of mutual information, we can obtain the following equation,
    \begin{equation}
    \label{mutual}
        \mathrm{I}(\mathrm{Z};\mathrm{T})=\mathbb H(\mathrm{Z}) - \mathbb H(\mathrm{Z}|\mathrm{T})\overset{(a)}{=}\mathbb H(\mathrm{Z}) - \mathbb H(\mathrm{Z},\mathrm{T})+\mathbb H(\mathrm{T}),
    \end{equation}
    where the equality $(a)$ holds due to the definition of conditional entropy, \textit{i.e.}, $\mathbb H(\mathrm{Z}|\mathrm{T})=\mathbb H(\mathrm{Z},\mathrm{T})-\mathbb H(\mathrm{T})$.

    Considering that the node representation $\mathrm{Z}=\mathrm{\tilde{H}}||\mathrm{\tilde{S}}$ used for downstream analysis is derived from the concatenation of $\mathrm{\tilde{H}}$ and $\mathrm{\tilde{S}}$, we have
    \begin{equation}
    \label{entropy}
        \begin{aligned}
            \mathbb H(\mathrm{Z})&=\mathbb H(\mathrm{\tilde{H}}|\mathrm{\tilde{S}})+\mathbb H(\mathrm{\tilde{S}}|\mathrm{\tilde{H}})+\mathrm{I}(\mathrm{\tilde{H}}; \mathrm{\tilde{S}}) \\
            &=\mathbb H(\mathrm{\tilde{H}}|\mathrm{\tilde{S}})+\mathbb H(\mathrm{\tilde{S}}|\mathrm{\tilde{H}})+\mathbb H(\mathrm{\tilde{H}})-\mathbb H(\mathrm{\tilde{H}}|\mathrm{\tilde{S}}) \\
            &\overset{(b)}{=}\mathbb H(\mathrm{\tilde{H}}|\mathrm{\tilde{S}})+\mathbb H(\mathrm{\tilde{S}}|\mathrm{\tilde{H}})+\mathbb H(\mathrm{\tilde{H}},\mathrm{\tilde{S}})-\mathbb H(\mathrm{\tilde{S}}|\mathrm{\tilde{H}})-\mathbb H(\mathrm{\tilde{H}}|\mathrm{\tilde{S}}) \\
            &=\mathbb H(\tilde{\mathrm{H}},\tilde{\mathrm{S}}),
        \end{aligned}
    \end{equation}
    where $\mathbb H(\mathrm{\tilde{H}}|\mathrm{\tilde{S}})$ and $\mathbb H(\mathrm{\tilde{S}}|\mathrm{\tilde{H}})$ indicate the unique information of $\mathrm{\tilde{H}}$ and $\mathrm{\tilde{S}}$. $\mathrm{I}(\mathrm{\tilde{H}}; \mathrm{\tilde{S}})$ is the shared information between $\mathrm{\tilde{H}}$ and $\mathrm{\tilde{S}}$. The equality $(b)$ holds because $\mathbb H(\tilde{\mathrm{H}})=\mathbb H(\tilde{\mathrm{H}},\tilde{\mathrm{S}})-\mathbb H(\tilde{\mathrm{S}}|\tilde{\mathrm{H}})$.
    By inserting Eq.\ref{entropy} into Eq.\ref{mutual}, we can obtain the following the equation,
    \begin{equation}
    \label{final_mutual_information}
        \begin{aligned}
            \mathrm{I}(\mathrm{Z};\mathrm{T})=\mathbb H(\tilde{\mathrm{H}},\tilde{\mathrm{S}})-\mathbb H(\tilde{\mathrm{H}},\tilde{\mathrm{S}},\mathrm{T})+\mathbb H(\mathrm{T}).
        \end{aligned}
    \end{equation}
    According to the definition of mutual information, we have 
    \begin{equation}
        \label{instance_information}
        \mathrm{I}(\tilde{\mathrm{H}};\mathrm{T})=\mathbb H(\mathrm{\tilde{H}})-\mathbb H(\tilde{\mathrm{H}}|\mathrm{T})=\mathbb H(\mathrm{\tilde{H}})-\mathbb H(\tilde{\mathrm{H}},\mathrm{T})+\mathbb H(\mathrm{T}).
    \end{equation}
    We subtract Eq.\ref{instance_information} from Eq.\ref{final_mutual_information}, yielding,
    \begin{equation}
    \label{relation}
        \begin{aligned}
            \mathrm{I}(\mathrm{Z};\mathrm{T})-\mathrm{I}(\tilde{\mathrm{H}};\mathrm{T})&=\mathbb H(\tilde{\mathrm{H}},\tilde{\mathrm{S}})-\mathbb H(\tilde{\mathrm{H}},\tilde{\mathrm{S}},\mathrm{T})-\mathbb H(\mathrm{\tilde{H}})+\mathbb H(\tilde{\mathrm{H}},\mathrm{T}) \\
            &=\underbrace{\mathbb H(\tilde{\mathrm{H}},\tilde{\mathrm{S}})-\mathbb H(\mathrm{\tilde{H}})}_{(c)}-\underbrace{(\mathbb H(\tilde{\mathrm{H}},\tilde{\mathrm{S}},\mathrm{T})-\mathbb H(\tilde{\mathrm{H}},\mathrm{T}))}_{(d)}\\
            &=\mathbb H(\tilde{\mathrm{S}}|\tilde{\mathrm{H}})-\mathbb H(\tilde{\mathrm{S}}|\tilde{\mathrm{H}},\mathrm{T})\\
            &\overset{(e)}{=}\mathrm{I}(\tilde{\mathrm{S}},\mathrm{T}|\tilde{\mathrm{H}})\overset{(f)}{\geq}0,
        \end{aligned}
    \end{equation}
    where the validity of equations $(c)$ and $(d)$ arises from the properties between conditional entropy and entropy. The equation $(e)$ holds due to the relationship between conditional mutual information and entropy. Moreover, the inequality $(f)$ holds because of the non-negativity property of mutual information. Next, we can naturally derive $\mathrm{I}(\mathrm{Z};\mathrm{T})-\mathrm{I}(\tilde{\mathrm{H}};\mathrm{T})\geq 0$. Similarly, following the same proof method, we can also conclude that $\mathrm{I}(\mathrm{Z};\mathrm{T})-\mathrm{I}(\tilde{\mathrm{S}};\mathrm{T})\geq 0$. $\mathrm{I}(\mathrm{Z};\mathrm{T})\geq\max{\{\mathrm{I}(\tilde{\mathrm{H}};\mathrm{T}),\mathrm{I}(\tilde{\mathrm{S}};\mathrm{T})\}}$ holds true. Therefore, we complete the proof of Theorem \ref{mutual_information}.
\end{proof}

\subsubsection{Proof of Corollary \ref{task_error}}
\begin{proof}
    To prove the Corollary \ref{task_error}, we first give the inequality between the Bayes error rate $\mathrm{P}(\mathrm{Z})$ and conditional entropy $\mathbb H(\mathrm{Y}|\mathrm{Z})$ given in previous work \cite{hellman1970probability} as follows:
    \begin{equation}
    \label{bayes_condition}
        \mathrm{P}(\mathrm{Z})\leq\frac{\mathbb H(\mathrm{Y}|\mathrm{Z})}{2},
    \end{equation}
    where $\mathrm{Y}$ denotes the class labels. Based on the definition of conditional entropy, we have,
    \begin{equation}
    \begin{aligned}
        &\mathbb H(\mathrm{Y}|\mathrm{Z})=\mathbb H(\mathrm{Y})-\mathrm{I}(\mathrm{Y};\mathrm{Z})\\
        &\mathbb H(\mathrm{Y}|\tilde{\mathrm{H}})=\mathbb H(\mathrm{Y})-\mathrm{I}(\mathrm{Y};\tilde{\mathrm{H}})
        \overset{(g)}{=}\mathbb H(\mathrm{Y})-\mathrm{I}(\mathrm{Y};\mathrm{Z})+\mathrm{I}(\tilde{\mathrm{S}},\mathrm{T}|\tilde{\mathrm{H}}),
    \end{aligned}
    \end{equation}
    where the equality $(g)$ holds due to Eq.\ref{relation}.
    
    Due to $\mathrm{I}(\tilde{\mathrm{S}},\mathrm{T}|\tilde{\mathrm{H}})\geq0$, hence $\mathbb H(\mathrm{Y}|\mathrm{Z})\geq \mathbb H(\mathrm{Y}|\tilde{\mathrm{H}})$. According to Inequality \ref{bayes_condition}, we can conclude that the maximum value of $\mathrm{P}(\mathrm{Z})$ is smaller than that of $\mathrm{P}(\tilde{\mathrm{H}})$. Therefore, $\mathrm{U}(\mathrm{P}(\mathrm{Z}))\leq\mathrm{U}(\mathrm{P}(\tilde{\mathrm{H}})$ is true. Similarly, we can obtain $\mathrm{U}(\mathrm{P}(\mathrm{Z}))\leq\mathrm{U}(\mathrm{P}(\tilde{\mathrm{S}})$. Combining the above results, we can prove that $\mathrm{U}(\mathrm{P}(\mathrm{Z})) \leq \min\{\mathrm{U}(\mathrm{P}(\mathrm{\tilde{H}})), \mathrm{U}(\mathrm{P}(\mathrm{\tilde{S}}))\}$. We complete the proof of Corollary \ref{task_error}.
\end{proof}

\subsubsection{Proof of Theorem \ref{upper_bound}}
\begin{proof}
To prove Theorem \ref{upper_bound}, we first give the following standard uniform deviation bound from \cite{bartlett2002rademacher}.
\begin{lemma}
\label{uniform_deviation}
    Let the sample $\{z_1,\cdots,z_N\}$ be drawn from a distribution over $\mathcal{Z}$. $\mathcal{H}$ is a class of functions on $\mathcal{Z}$. For $\delta>0$, we have that with a probability at $1-\delta$ over the draw of the sample,
    \begin{equation}
    \underset{\text{h}\in\mathcal{H}}{\text{sup}}|\mathbb E_{\hat{\mathbb P}}h(z)-\mathbb E_\mathbb Ph(z)|\leq 2\mathcal{R}(\mathcal{H})+\sqrt{\frac{\log(1/\delta)}{N}},
    \end{equation}
    where $\hat{\mathbb P}$ and $\mathbb P$ denote the empirical and expected data distribution. $\mathcal{R}$ is the Rademacher complexity of the function class $\mathcal{H}$ with respective to samples.
\end{lemma}

According to Lemma \ref{uniform_deviation}, and the definitions of $\hat{\mathsf{R}}$, $\mathsf{R}$, and $\mathcal{F}_\gamma$ given in Section \ref{theoretical_analysis}, we have,
\begin{equation}
    |\hat{\mathsf{R}}-\mathsf{R}|\leq2\mathcal{R}(\mathcal{F}_\gamma)+\sqrt{\frac{\log(1/\delta)}{2n}}.
\end{equation}
Let $\zeta_1,\cdots,\zeta_n$ represent uniform random Rademacher variables taking values from $[-1,1]$. Based on the definition of Rademacher complexity, we can bound the empirical one as follows:
\begin{equation}
\label{gap}
\begin{aligned}
    &\hat{\mathcal{R}}(\mathcal{F}_\gamma)=\mathbb E_\zeta\underset{f\in\mathcal{F}_\gamma}{\text{sup}}\frac{1}{n}\sum_{i=1}^n\zeta_if(\mathrm{Z}_i)=\mathbb E_{\zeta}\underset{\theta\Sigma\theta^\top\leq\gamma}{\text{sup}}\frac{1}{n}\sum_{i=1}^n\zeta_i\theta^\top\mathrm{Z}_i\\
    &\leq\frac{1}{n}\mathbb E_\zeta\underset{\theta\Sigma\theta^\top\leq\gamma}{\text{sup}}\left|\left|\Sigma^{1/2}\theta\right|\right|\left|\left|\sum_{i=1}^n\zeta_i\Sigma^{\dag/2}\mathrm{Z}_i\right|\right|\\
    &\leq\frac{\sqrt{\gamma}}{n}\mathbb E_\zeta\sqrt{\sum_{i=1}^n\zeta_i(\Sigma^{\dag/2}\mathrm{Z}_i)^\top\Sigma^{\dag/2}\mathrm{Z}_i}\\
    &\leq\frac{\sqrt{\gamma}}{n}\sqrt{\mathbb E_\zeta\sum_{i=1}^n\zeta_i(\Sigma^{\dag/2}\mathrm{Z}_i)^\top\Sigma^{\dag/2}\mathrm{Z}_i}\\
    &=\frac{\sqrt{\gamma}}{n}\sqrt{\sum_{i=1}^n(\Sigma^{\dag/2}\mathrm{Z}_i)^\top\Sigma^{\dag/2}\mathrm{Z}_i}\\
    &=\frac{\sqrt{\gamma}}{n}\sqrt{\sum_{i=1}^n(\mathrm{Z}_i)^\top\Sigma^{\dag}\mathrm{Z}_i},
\end{aligned}
\end{equation}
where $\Sigma^\dag$ is the generalized inverse of $\Sigma$.
Next, we can bound the Rademacher complexity by taking expectations as follows:
\begin{equation}
\label{rc}
\begin{aligned}
    &\mathcal{R}(\mathcal{F}_\gamma)=\mathbb E_\mathrm{Z}\hat{\mathcal{R}}(\mathcal{F}_\gamma)\leq\mathbb E_\mathrm{Z}\frac{\sqrt{\gamma}}{n}\sqrt{\sum_{i=1}^n(\mathrm{Z}_i)^\top\Sigma^{\dag}\mathrm{Z}_i}\\ 
    &\leq\frac{\sqrt{\gamma}}{n}\sqrt{\sum_{i=1}^n\mathbb E_\mathrm{Z}(\mathrm{Z}_i)^\top\Sigma^{\dag}\mathrm{Z}_i}=\frac{\sqrt{\gamma}}{n}\sqrt{\sum_{i=1}^n\Sigma^{\dag}\mathbb E_\mathrm{Z}(\mathrm{Z}_i)^\top\mathrm{Z}_i}\\
    &=\frac{\sqrt{\gamma}}{n}\sqrt{\sum_{i=1}^n\sum_{l,s}(\Sigma^{\dag})_{ls}\mathbb E_\mathrm{Z}(\mathrm{Z}_i)_l(\mathrm{Z}_i)_s}=\frac{\sqrt{\gamma}}{n}\sqrt{\sum_{i=1}^n\sum_{l,s}(\Sigma^{\dag})_{ls}(\Sigma)_{ls}}\\
    &=\frac{\sqrt{\gamma}}{n}\sqrt{\sum_{i=1}^n\text{tr}(\Sigma\Sigma^\dag)}=\frac{\sqrt{\gamma}}{n}\sqrt{\sum_{i=1}^n\text{rank}(\Sigma)}=\sqrt{\frac{\gamma\cdot\text{rank}(\Sigma)}{n}}.
\end{aligned}
\end{equation}
When inserting Eq.\ref{rc} into Inequality \ref{gap}, we have,
\begin{equation}
    |\hat{\mathsf{R}}-\mathsf{R}|\leq2\sqrt{\frac{\gamma\cdot\text{rank}(\Sigma)}{n}}+\sqrt{\frac{\log(1/\delta)}{2n}}.
\end{equation}
When replacing $n=NK$, we can obtain the desired results. Thus, we complete the proof of Theorem \ref{upper_bound}.
\end{proof}

\subsection{Details of Datasets}
\label{dataset_detail}
We provide the details of the used datasets in the following.

\noindent \textbf{CoraFull} \cite{bojchevski2017deep}: It is a citation network. Each node represents a paper, and if one paper cites another, there is an edge between them. Nodes are labeled based on the topics of the papers. For this dataset, we use 25/20/25 node class splits for meta-training/meta-validation/meta-testing.

\noindent \textbf{Coauthor-CS} \cite{shchur2018pitfalls}: It is a co-authorship graph where nodes mean authors and edges mean co-authorship between authors. For this dataset, we use 5/5/5 node classes for meta-training/meta-validation/\\meta-testing.

\noindent \textbf{Cora} \cite{yang2016revisiting}: It is also a citation network where nodes represent papers and edges represent citation relationships. For this dataset, we use 3/2/2 class splits for meta-training/meta-validation/meta-testing.

\noindent \textbf{WikiCS} \cite{shchur2018pitfalls}: It is a graph dataset obtained from Wikipedia, where nodes represent computer science articles, and edges are constructed based on hyperlinks. We perform 4/3/3 class splits for meta-training/\\meta-validation/meta-testing.

\noindent \textbf{ML} \cite{bojchevski2017deep}: It is a dataset containing only machine learning papers, where nodes represent papers, and an edge is constructed between two papers if there is a citation relationship between them. We use 3/2/2 node classes for meta-training/meta-validation/meta-testing.

\noindent \textbf{CiteSeer} \cite{yang2016revisiting}: It is a citation graph where nodes are scientific publications and edges are citation relationships. We use 2/2/2 node classes for meta-training/meta-validation/meta-testing.

\noindent \textbf{ogbn-arxiv} \cite{hu2020open}: It is a citation network of computer science papers, where nodes represent the arXiv papers, and edges are constructed between papers if there exists a citation between them. Node labels are assigned based on 40 CS subject areas in arXiv. For this dataset, we adopted 20/10/10 class splits.

\noindent \textbf{ogbn-products} \cite{hu2020open}: It is an Amazon product co-purchasing network, where nodes represent products, and an edge is constructed between two products if they are simultaneously purchased. Node labels represent the category of the items. We used 23/12/12 class splits for this dataset.

\subsection{Details of Baselines}
\label{baseline_descrip}
\subsubsection{Graph embedding methods}
\ 

\noindent \textbf{DeepWalk} \cite{perozzi2014deepwalk}: It performs random walk based on the word2vec algorithm on the graph and obtains low dimensional node embeddings. 

\noindent \textbf{node2vec} \cite{grover2016node2vec}: It further extends DeepWalk by conducting biased random walk such as depth-fist search or breadth-first search on the graph.

\noindent \textbf{GCN} \cite{kipf2016semi}: It applies a first-order Chebyshev approximation graph filter on the graph to obtain hidden node embeddings for downstream task analysis.

\noindent \textbf{SGC} \cite{wu2019simplifying}: It further simplifies the GCN by removing the non-linear activation and collapsing the weights.

\subsubsection{Traditional meta learning methods}
\ 

\noindent \textbf{ProtoNet} \cite{snell2017prototypical}: It aims to learn a metric space and compute the similarity between query samples and each prototype obtained from support samples to predict their categories.

\noindent \textbf{MAML} \cite{finn2017model}: It performs one or more gradient steps on the parameters to allow the meta-learner to obtain well-initialized parameters, enabling rapid adaptation to downstream tasks with limited annotated data.

\subsubsection{Graph meta learning methods}
\ 

\noindent \textbf{Meta-GNN} \cite{zhou2019meta}: It seamlessly integrates MAML and GNN in a straightforward manner, leveraging the MAML framework to acquire useful prior knowledge from previous tasks, enabling rapid adaptation to new tasks.

\noindent \textbf{GPN} \cite{ding2020graph}: It employs a graph encoder and evaluator based on ProtoNet to learn node embeddings and assess the importance of these nodes, while also assigning new samples to the class closest to them.

\noindent \textbf{G-Meta} \cite{huang2020graph}: It constructs a separate subgraph for each node, transmitting specific node information while leveraging meta-gradients to learn transferable knowledge.

\noindent \textbf{TENT} \cite{wang2022task}: It proposes an adaptive framework comprising node-level, class-level, and task-level components, aiming to reduce the generalization gap between meta-training and meta-testing, and mitigate the impact of diverse task variations on model performance.

\noindent \textbf{Meta-GPS} \cite{liu2022few}: It cleverly introduces prototype-based parameter initialization, scaling, and shifting transformations, thereby facilitating better learning of transferable meta-knowledge within the MAML framework and faster adaptation to new tasks.

\noindent \textbf{X-FNC} \cite{wang2023few}: It first conducts label propagation based on Poisson learning to obtain rich pseudo-labeled nodes, and then filters out irrelevant information through node classification and an information bottleneck-based method.

\noindent \textbf{TEG} \cite{kim2023task}: It designs a task-equivariant graph few-shot learning framework, leveraging equivariant neural networks to learn adaptive strategies for specific tasks, aimed at capturing task-inductive biases.

\noindent \textbf{COSMIC} \cite{wang2023contrastive}: It proposes a contrastive meta-learning framework, which first explicitly aligns node embeddings within each episode through a two-step optimization process.

\noindent \textbf{TLP} \cite{tan2022transductive}: It introduces the concept of transductive linear probing, initially pretraining graph encoders through graph contrastive learning, and then applying it to obtain node embeddings during the meta-testing phase.